\documentclass[letterpaper, 10 pt, conference]{ieeeconf}
\usepackage{times}
\usepackage[pdftex]{graphicx}
\usepackage{amsmath,amssymb,amsopn,amstext,amsfonts}
\usepackage{cancel}
\usepackage[space]{cite}
\usepackage{pdfsync}
\usepackage{balance}
\usepackage{color}
\usepackage{mathtools}
\usepackage{algpseudocode}
\usepackage{algorithm} %\usepackage[ruled,vlined,linesnumbered]{algorithm2e}
\usepackage{bm}

\usepackage{diagbox}
\usepackage{float}
\usepackage{epstopdf}
\usepackage{pifont}
\usepackage{fixltx2e}
\usepackage{amsmath}
\usepackage{multirow}
\usepackage{url}
\usepackage{verbatim}
\usepackage[linkcolor=black,citecolor=black,urlcolor=black,colorlinks=true]{hyperref}
\usepackage{booktabs}
\usepackage{graphicx}
\usepackage{subcaption}

\algnewcommand\algorithmicforeach{\textbf{for each}}
\algdef{S}[FOR]{ForEach}[1]{\algorithmicforeach\ #1\ \algorithmicdo}
\newcommand{\tp}{^{\mathsf{T}}}

\graphicspath{{figures/}}
\DeclareGraphicsExtensions{.png,.jpg,.eps,.pdf}
\IEEEoverridecommandlockouts
\title{\LARGE \bf Integrating Fast Regional Optimization into Sampling-based Kinodynamic Planning for Multirotor Flight}
\author
{Hongkai Ye$^{1, 2}$, Tianyu Liu$^{3}$, Chao Xu$^{1, 2}$ and Fei Gao$^{1, 2}$
\thanks{
$^{1}$State Key Laboratory of Industrial Control Technology, Zhejiang University, Hangzhou 310027, China.}
\thanks{
$^{2}$Huzhou Institute of Zhejiang University, Huzhou 313000, China.}
\thanks{
$^{3}$Department of Mechanical Engineering, The University of Hong Kong, Hong Kong, China.}
\thanks{\tt Email: hkye@zju.edu.cn, tianyu@connect.hku.hk, cxu@zju.edu.cn and fgaoaa@zju.edu.cn}
}

\begin{document}
\maketitle
\thispagestyle{empty}
\pagestyle{empty}

\begin{abstract}
For real-time multirotor kinodynamic motion planning, the efficiency of sampling-based methods is usually hindered by difficult-to-sample homotopy classes like narrow passages.
In this paper, we address this issue by a hybrid scheme.
We firstly propose a fast regional optimizer exploiting the information of local environments and then integrate it into a global sampling process to ensure faster convergence.
The incorporation of local optimization on different sampling-based methods shows significantly improved success rates and less planning time in various types of challenging environments.
We also present a refinement module that fully investigates the resulting trajectory of the global sampling and greatly improves its smoothness with negligible computation effort.
Benchmark results illustrate that compared to the state-of-the-art ones, our proposed method can better exploit a previous trajectory.
The planning methods are applied to generate trajectories for a simulated quadrotor system, and its capability is validated in real-time applications.

\end{abstract}

\section{Introduction}
\label{sec:introduction}

Multirotors can conduct agile maneuvers like catching a ball in the air\cite{mueller2015computationally}, flying through narrow passages\cite{liu2018ral}, or even acrobatics\cite{kaufmann2020RSS}.
To enable such mobility and to prevent making motion plans that can not be fulfilled, the system dynamics, as well as the control and state saturation constraints, have to be considered when planning trajectories, which usually relate to kinodynamic motion planning\cite{Donald1993kinodynamic}.

With the capability of globally reasoning for an optimal trajectory in exploring the entire nonconvex solution space, some kinodynamic variants of path planning methods have been successfully applied to solve the problem with various system settings.
Search-based ones\cite{liu2017iros, liu2018ral, likhachev2009planning} discretize the control space and expand states with dynamics integration. Although an optimal solution is most likely guaranteed with carefully designed and admissible heuristics, the discretization introduces a compromise on solution quality and solving time. Finely discretized controls are prone to find a good solution but may require untrackable solving time.
This curse of dimensionality makes it unsuitable for applications where real-time planning is required.

Many sampling-based variants\cite{Schmerling2015drift, Schmerling2015driftless, webb2013kinodynamic}, on the other hand, are asymptotically optimal and hold the anytime property, meaning that an initial solution, whether coarse or not, can be first obtained quickly and improved with an extra computation budget. They seek to identify feasible motions in the continuous solution space by drawing discrete samples and build connections between them.
Solving Boundary Value Problems (BVPs) is almost inevitable for building these connections, which is difficult, especially for nonlinear dynamics with a non-differentiable objective and complex constraints.
Our previous work~\cite{Hongkai2021tgk} eases this problem by relaxing constraints with linear models of multirotors. Still, it is not efficient to explore the whole solution space if there exist difficult-to-sample homotopy classes like narrow passages, which can cost great effort for a standalone sampling-based kinodynamic planner to sample through.
Local trajectory optimization techniques can be great compensations for these situations since they prioritize exploiting domain information to explore the local environment.
As a result, integrating regional optimization (RO) into global exploration is embraced by many works~\cite{Choudhury2016Regionally, Kim2018Dancing, Kim2019Volumetric} and become the trends to deal with such problems. Empirically, this scheme's overall effect is dominated by the manner of integration and the solving time of the local optimizer.

\begin{figure}[t]
\centering
\begin{subfigure}{0.9\linewidth}
	\includegraphics[width=1\linewidth]{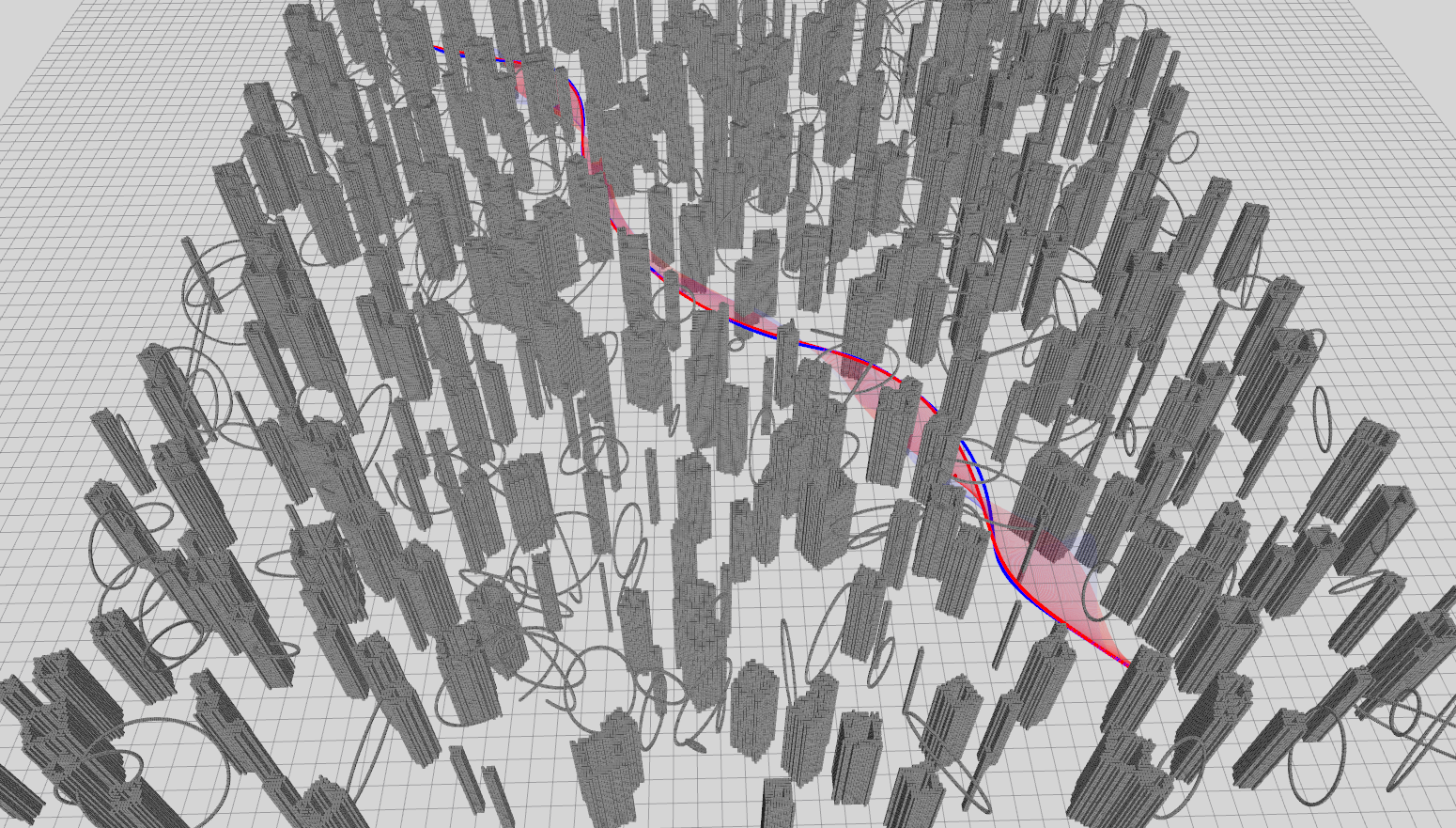}	
	\caption{The simulation environment}
	\label{subfig:regional_ap}
\end{subfigure}
\begin{subfigure}{0.9\linewidth}
	\includegraphics[width=1\linewidth]{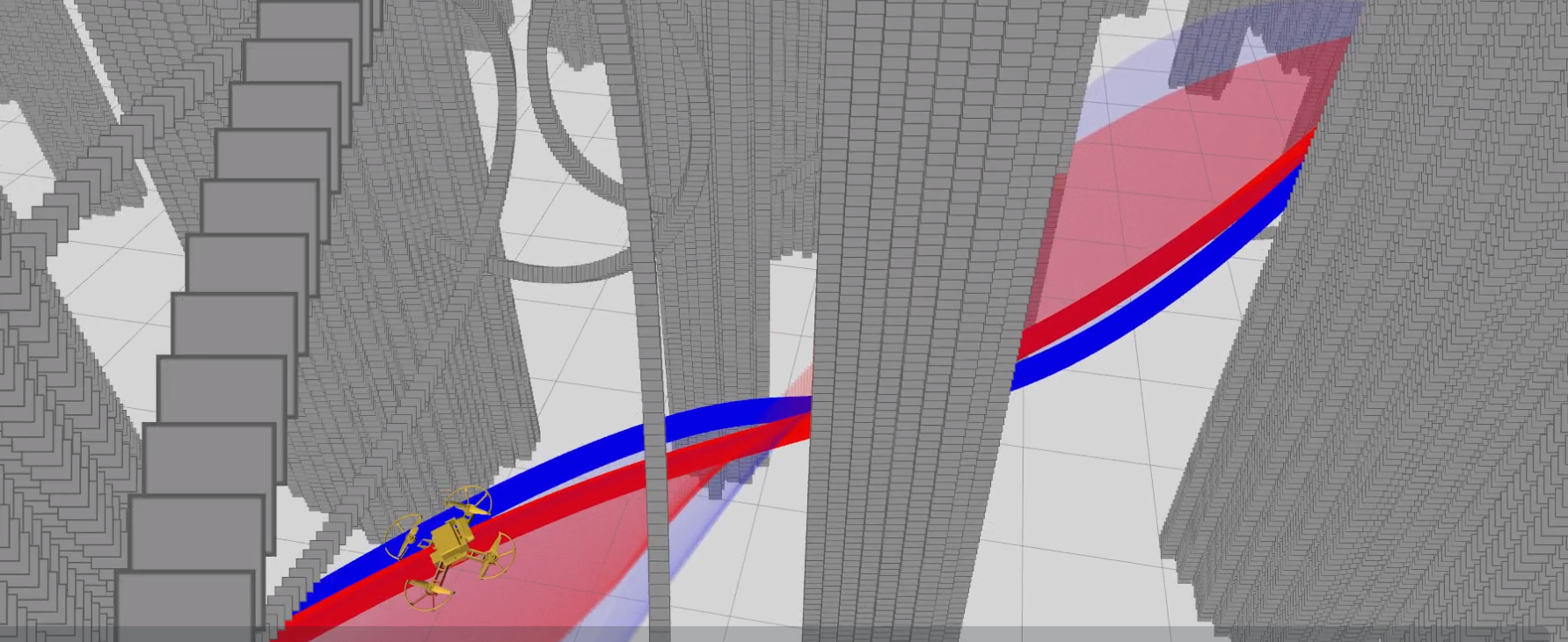}
	\caption{The transparent colored lines indicate the accelerations.}
	\label{subfig:backend_ap}
\end{subfigure}
\captionsetup{font={small}}
\caption{A multirotor flying in simulation. The blue and red curves indicate the trajectories of the proposed front-end and back-end, respectively. The smoothness is improved by the back-end.}
\label{fig:simulation}
\vspace{-0.7cm}
\end{figure}

The resulting trajectory of the aforementioned global kinodynamic planning, though satisfies all the constraints and settles in a proper homotopy class, may still not be smooth enough with unpleasant motions to be directly fed to a vehicle, which lefts space for refining.
Many works ~\cite{dolgov2010path, boyu2019ral, Gao2018Online} adopt a lightweight refinement module to preserve and improve some properties of the previous result, obtaining much better outcomes. This two-stage scheme is commonly known as the front-end $\&$ back-end framework of kinodynamic motion planning. Nevertheless, it remains open to making full use of the front-end results while retaining efficiency, effectiveness, and high success rate.

In this paper, as an enhancement of our previous work~\cite{Hongkai2021tgk},
we focus on the gaps mentioned above that 1) narrow passages are difficult to address in the global sampling process, and 2) front-end assets are not fully exploited in the back-end refining. We present a kinodynamic planning method and framework that meets real-time requirements for multirotor flight.
Utilizing the global reasoning capability and the anytime property of some sampling-based kinodynamic planners, a variant of the kinodynamic RRT* (kRRT*)\cite{webb2013kinodynamic, Hongkai2021tgk} is adopted as the front-end to conduct the global search. A notable difference is that we integrate a fast regional trajectory optimizer into the global reasoning process, greatly accelerating to find an initial solution with a higher success rate. The regional optimization is formulated as a sequence of quadratic programming (QP) with a closed-form solution for each iteration, which guarantees fast solution time and avoids the requirements of calculating any gradients.
Inheriting the ideas from our regional trajectory optimization and benefiting from the front-end trajectory, we also develop a lightweight and practical back-end refinement module with extensions to consider the obstacle clearance in some degree. This refinement is extremely fast and remarkably improves the success rate.

The main contributions of this paper are:
\begin{enumerate}
\item Developing a fast trajectory optimization method as a sequence of QP based on \cite{Hongkai2021tgk} and \cite{bry2015aggressive}, which obtains feasible solutions with much fewer iterations by taking obstacle clearance into account while a closed-form solution is preserved for each iteration.
\item Integrating the method as a regional trajectory optimizer into an RRT*-based kinodynamic planner\cite{Hongkai2021tgk}. Planning under this framework improves the initial solution quality, success rate, and convergence rate of the global search.
\item Using the optimization method as a lightweight trajectory refinement back-end, which exploits the kinodynamic global planning assets and improves the resulting trajectory in smoothness and obstacle clearance efficiently.
\item Applying the proposed planning methods to trajectory generation of a simulated quadrotor system (shown in Fig.~\ref{fig:simulation}), presenting extensive benchmarks and simulated experimental validations, and releasing source code for the reference of the community. \footnote{Code at \url{https://github.com/kyleYehh/kino_sampling_with_regional_opti} associated with a simulation video.}
\end{enumerate}

\section{Related Works}
%\subsection{Multirotors Kinodynamic planning}
%, whereas sparsely discritized controls harm the resolution completeness and may fail to find a solution.
%Some works\cite{Boeuf2014Planning, rauscher2020efficient} design specfic control patterns to avoid it but they only apply for time optimal trajectories with integrator chain dynamics, and
%RRT-like relate to order of sampling sequence

\subsection{Planning with Regional Optimization}
Most standalone search-based and sampling-based global planning methods solely focus on exploring the entire solution space to bring an optimal answer. Local domain information is disregarded, and as a result, they suffer from low efficiency when narrow passages present.
Regional optimizers, on the other hand, prioritize investigating a limited area thus are better with narrow spaces.
Choudhury et al.~\cite{Choudhury2016Regionally} combine gradient-based local optimizers like CHOMP\cite{zucker2013chomp} with their global sampling process adopted from BIT*~\cite{Gammell2015bit} and achieve faster access to an initial solution as well as a higher convergence rate in high dimensions. However, when few difficult-to-sample homotopy class exists, the relatively heavy optimization can rather hinder the global process.
Another disadvantage is that extra efforts are required to calculate the obstacle distance gradients a priori.
Kim et al.~\cite{Kim2018Dancing, Kim2019Volumetric} follow similar schemes and develop a PRM-style~\cite{karaman2011} sparse graph to explore different homotopy classes. They instead get obstacle gradients with empirical collisions found during the execution. Though avoid prior computation, it is uneasy to obtain accurate gradients and thus impedes the convergence.
Unlike all these works that focus on path planning problems, we instead present trajectory planning for multirotor kinodynamic systems, incorporating an efficient regional optimizer that requires no gradient information or any prior computation.

\subsection{Exploiting Front-end Result}
Though all the constraints are satisfied, the trajectory obtained by the global reasoning part may lack smoothness due to the limited time budget. A refinement module is usually followed to improve it.
Some works~\cite{boyu2019ral, Gao2018Online, liu2017ral} re-parameterize their front-end result in the back-end and form optimization problems with soft constraints, where a distance field or/and free corridors are obliged to provide obstacle clearance.
Liu et al.~\cite{liu2017iros} solve a unconstrainted QP with intermediate waypoints and time allocation obtained from their search-based kinodynamic front-end and check feasibility afterward. They do not need to build any time-consuming fields or corridors but require the intermediate waypoints fixed to provide collision-free information in the environments.
Our previous work~\cite{Hongkai2021tgk} relax this constraint and instead formulate a \textit{Homotopy Cost} to force every point in the optimized trajectory to be close to the original collision-free one. However, no obstacle information is considered, and thus it takes many iterations to get a feasible result or even fail in complex environments. This work addresses it by further incorporating collision penalties found during each iteration on specific parts of the optimized trajectory.

\begin{algorithm}[t]
\caption{\\Kinodynamic RRT* with Regional Optimization}
\label{alg:kino_rrt_star}
\begin{algorithmic}[1]
\State \textbf{Notation}: Environment $\mathcal{E}$, Tree $\mathcal{T}$, State $\mathbf{x}$
\State Initialize: $\mathcal{T} \leftarrow \emptyset \cup \{\mathbf{x}_{init}\}$
\For{$i = 1$ to $n$}
    \State $\mathbf{x}_{random} \leftarrow$ \textbf{Sample}($\mathcal{E}$)
    \State $\mathcal{X}_{backward} \leftarrow$ \textbf{BackwardNear}($\mathcal{T}$, $\mathbf{x}_{random}$)
    \State $\mathbf{x}_{min} \leftarrow$ \textbf{ChooseParent}($\mathcal{X}_{backward}$, $\mathbf{x}_{random}$)
    \State $\mathcal{T} \leftarrow \mathcal{T} \cup \{\mathbf{x}_{min}, \mathbf{x}_{random}\}$
    \If{\textbf{TryConnectGoal}($\mathbf{x}_{random}$, $\mathbf{x}_{goal}$)}
    	\State break
    \EndIf
    \State $\mathcal{X}_{forward} \leftarrow$ \textbf{ForwardNear}($\mathcal{T}$, $\mathbf{x}_{random}$)
    \State \textbf{Rewire}($\mathcal{T}$, $\mathcal{X}_{forward}$)
\EndFor
\State \Return $\mathcal{T}$
\end{algorithmic}
\end{algorithm}

\section{Methodology}
We begin by presenting the problem definition, and then introduce the modified kRRT* framework, highlighting the integrated differences, and then present the quadratic formulation and closed-form solution of the polynomial trajectory optimizer, and finally, the refinement back-end.

\subsection{Preliminaries}
With the help of the multirotor's differential flatness property\cite{MelKum1105}, it allows us to use a linear model to represent its dynamics with four flat outputs $p_x, p_y, p_z$ (position in each axis), $\psi$ (yaw) and their derivatives for the state variables.
%The original states and inputs can be recovered from the flat outputs, and any smooth trajectory can be tracked with reasonably bounded derivatives of the flat output space.
To impose continuity in at least acceleration, we use the triple integrator model with jerk as the input:

\begin{equation}
\label{equ:dynamics}
\mathbf{\dot{x}}(t)=\mathbf{A}\mathbf{x}(t)+\mathbf{B}\mathbf{u}(t),
\end{equation}
\begin{equation}
\begin{split}
&\mathbf{x}(t)=
\begin{bmatrix}
\mathbf{p}(t) \\
\mathbf{\dot{p}}(t) \\
\mathbf{\ddot{p}}(t)
\end{bmatrix}, \quad
\mathbf{A}=
\begin{bmatrix}
\mathbf{0} & \mathbf{I} \\
\mathbf{0} & \mathbf{0}
\end{bmatrix}, \quad
\mathbf{B}=
\begin{bmatrix}
\mathbf{0} \\
\mathbf{I}
\end{bmatrix}, \\[1ex]
&\mathbf{u}(t)=\mathbf{\dddot{p}}(t), \quad
\mathbf{p}(t) =
\begin{bmatrix}
p_x(t),\ p_y(t),\ p_z(t)
\end{bmatrix}\tp.
\end{split}
\end{equation}

Following\cite{Hongkai2021tgk, liu2017iros, boyu2019ral}, we search for a trajectory starting from an initial state $\mathbf{x}_{init}$ to a goal state region $\mathcal{X}_{goal}$ that is optimal in a tradeoff between time and energy. The trajectory planning problem is formulated as follows:

\begin{equation}
\label{equ:problem_form}
\begin{split}
\min_{\mathbf{x}(t)} \mathcal{J}&=\int_{0}^{\tau}(\rho+\frac{1}{2}\mathbf{u}(t)\tp\mathbf{u}(t))dt \\[1ex]
s.t. \quad & \mathbf{A}\mathbf{x}(t)+\mathbf{B}\mathbf{u}(t) - \mathbf{\dot{x}}(t)=\mathbf{0}, \\
&\mathbf{x}(0)=\mathbf{x}_{init}, \
\mathbf{x}(\tau) \in \mathcal{X}_{goal}, \\
&\forall t \in [0, \tau],\
\mathbf{x}(t) \in \mathcal{X}^{free}, \
\mathbf{u}(t) \in \mathcal{U}^{free},
\end{split}
\end{equation}
where $\tau$ is the time duration of the trajectory and $\rho$ the tradeoff weight to penalize time against energy. $ \mathcal{X}^{free}$ stands for states that are obstacle-free and satisfy derivative constraints. $\mathcal{U}^{free}$ indicates feasible controls.

Polynomial splines are widely used to express time and energy optimal trajectories in continuous-time motion planning owing to their strong manifestations with simple parameterization. We use polynomials as trajectory bases in both front-end and back-end, and the final solution trajectory is expressed as piecewise polynomials.
For differential flat systems, each flat output dimension can be decoupled and planed, respectively.
For each dimention (yaw planning is excluded in this paper), consider an $(n+1)$-order, $m$-piece spline with its $i^{th}$ segment an $n$-degree polynomial
$p_i(t)=\mathbf{c}_i\tp\mathbf{t}, t \in [0,T_i]$,
where $\mathbf{c}_i \in \mathbb{R}^{n+1}$ is the coefficient vector of the $i^{th}$ segment, $\mathbf{t}=(1,t,t^2,...,t^n)\tp$ the natrual basis, and $T_i$ the time duration.

\begin{algorithm}[t]
\caption{\textbf{ChooseParent}($\mathcal{X}$, $\mathbf{x}$)}
\label{alg:ChooseParent}
\begin{algorithmic}[1]
\State \textbf{Notation}: Trajectory Edge $\mathbf{e}$
\State $\mathbf{x}_{min} \leftarrow null,\ min\_cost \leftarrow 0$
\For{$\mathbf{x}_{parent}$ in $\mathcal{X}$}
    \State $\mathbf{e} \leftarrow$ \textbf{StateTransit}($\mathbf{x}_{parent}$, $\mathbf{x}$)
    \If {\textbf{ConsSatis}($\mathbf{e}$) $\bigwedge$ $\textbf{Cost}(\mathbf{e}) < min\_cost$}
    	\State $\mathbf{x}_{min} \leftarrow \mathbf{x}_{parent},\ min\_cost \leftarrow \textbf{Cost}(\mathbf{e})$
    \Else
    	\If {\textbf{NeedOptimize}($\mathbf{e}$)}
    		\State $\mathbf{e} \leftarrow$ \textbf{RegionalOptimize}($\mathbf{x}_{parent}$, $\mathbf{x}$)
    		\If {\textbf{ConsSatis}($\mathbf{e}$) $\bigwedge$ $\textbf{Cost}(\mathbf{e}) < min\_cost$}
    			\State $\mathbf{x}_{min} \leftarrow \mathbf{x}_{parent},\ min\_cost \leftarrow \textbf{Cost}(\mathbf{e})$
    		\EndIf
    	\EndIf
    \EndIf
\EndFor
\State \Return $\mathbf{x}_{min}$
\end{algorithmic}
\end{algorithm}

\subsection{kRRT* with Regional Optimization}

The main workflow of the modified kRRT* is described in Alg.~\ref{alg:kino_rrt_star}, where a trajectory tree is grown from the initial state towards the goal state.
In \textbf{ChooseParent()} and \textbf{Rewire()}, the \textbf{StateTransit()} that builds a connecting trajectory between two states acts as a fundamental unit, and requires the solving of a BVP described with Equ.~\ref{equ:problem_form}. It is commonly known as the steer function in most RRT-like planners.
Since the function is likely to be called several times for each drawn sample, it is better to be fast to compute so that the solution space can be explored more thoroughly with more samples within a given time budget.

Building on \cite{Hongkai2021tgk}, we temporarily assume the state and control unbounded and solve it by Pontryagin Maximum Principle, obtaining the optimal transition time $\tau^*$ and deriving the unconstrained optimal transition trajectory as a $5^{th}$-degree polynomial in several microseconds. This piece of trajectory is then checked for constraints feasibility.
In \cite{Hongkai2021tgk}, the trajectory merely is aborted if any violation occurs. In this paper, however, we reuse the unqualified trajectory by regional optimization to improve the efficiency. If the derivative constraints are violated, we increase $\tau^*$ and recompute the polynomial coefficients until they are not. If the trajectory collides with any obstacle, it is checked for conditional regional optimization, as is presented in the next sections.

\subsection{Quadratic Objective Formulation}
\label{sec:ObjectiveFormulation}
Being an integrated part of the frequently called steer function, the optimization process needs to be computationally fast as well.
Investigating the piece of trajectory that is checked collided, it has a reasonable time duration, and we know where the collision occurs. It is desired to deform the trajectory to a collision-free one in its nearby free solution space with as little effort.
Enhancing the works in~\cite{bry2015aggressive, Hongkai2021tgk} where the objective is in quadratic form, and the matrix of the quadratic term is positive definite (PD) such that closed-form solutions are available, we further add an objective term of collision in aware of the obstacle areas. The single piece of trajectory is uniformly divided by time into $j$ pieces of the same degree to bring in more freedom for the following optimization.

For each axis out of $x$, $y$, and $z$, the quadratic objective consists of three terms:
\subsubsection{Smoothness Cost}
$J_s$ is formulated as the time integral of the squared jerk of the trajectory:
\begin{equation}
\begin{aligned}
J_s=&\int_0^T[p^{(3)}(t)]^2dt \\
%=&\sum_{i=1}^{j} \int_0^{T_i}[p_{i}^{(3)}(t)]^2dt \\
=&\sum_{i=1}^{j} \mathbf{c}_{i}\tp \int_0^{T_i}\mathbf{t}^{(3)} (\mathbf{t}^{(3)})\tp dt \ \mathbf{c}_{i}
%=&\sum_{i=1}^{m} \mathbf{c}_{i}\tp \mathbf{Q}_{s,i} \ \mathbf{c}_{i} \\
=\mathbf{c}\tp \mathbf{Q}_s \mathbf{c},
\end{aligned}
\end{equation}
where $T=T_1+T_2+\cdots+T_j$ is the total duration of the trajectory and $T_i$ the duration of one divided piece.
$\mathbf{c}\tp=[\mathbf{c}_{1}\tp, \mathbf{c}_{2}\tp, \cdots, \mathbf{c}_{j}\tp]$ is the coefficient vector of the $j$ segments.

\subsubsection{Resemblance Cost}
$J_r$ is formulated as the integration over the squared difference between positions of the optimized trajectory and the originally divided trajectory $p^*(t)$:
\begin{equation}
\begin{aligned}
J_r=&\int_0^T[p(t)-p^*(t)]^2dt \\
%=&\sum_{i=1}^{j} \int_0^{T_i}[p_{i}(t)-p_{i}^*(t)]^2dt \\
=&\sum_{i=1}^{j} (\mathbf{c}_{i}-\mathbf{c}_{i}^*)\tp \int_0^{T_i}\mathbf{t} \mathbf{t}\tp dt \ (\mathbf{c}_{i}-\mathbf{c}_{i}^*) \\
%=&\sum_{i=1}^{j} (\mathbf{c}_{i}-\mathbf{c}_{i}^*)\tp \mathbf{Q}_{r,i} \ (\mathbf{c}_{i}-\mathbf{c}_{i}^*) \\
=&(\mathbf{c}-\mathbf{c}^*)\tp \mathbf{Q}_r (\mathbf{c}-\mathbf{c}^*).
\end{aligned}
\end{equation}

This term drives the optimized trajectory to be close in position to the original one, and thus the collision-free parts are likely to remain feasible.

\begin{algorithm}[t]
\caption{\textbf{RegionalOptimize}($\mathbf{x_1}$, $\mathbf{x_2}$)}
\label{alg:RegionalOptimize}
\begin{algorithmic}[1]
\State $\mathbf{e} \leftarrow null$
\While{$iter\_num < max\_num$}
	\State \textbf{Adjust}()
    \State $\mathbf{e}_{temp} \leftarrow$ \textbf{ClosedFormSolve}()
    \If {\textbf{CheckFeasible}($\mathbf{e}_{temp}$, $\mathcal{E}$)}
		\State $\mathbf{e} \leftarrow \mathbf{e}_{temp}$
        \State break
    \Else
    	\State \textbf{Adjust}()
    \EndIf
\EndWhile
\State \Return $\mathbf{e}$
\end{algorithmic}
\end{algorithm}

\subsubsection{Collision Cost}
$J_c$ is formulated similar to the resemblance term. The difference is that the substracted trajectory $p^*(t)$ is some selected attracting points providing dragging force to draw the collided part to nearby collision-free areas.
\begin{equation}
\begin{aligned}
J_c=&\sum_{ap \in APs} \int_{t_{s,ap}}^{t_{e,ap}}[p(t)-p_{ap}(t)]^2dt \\
=&\sum_{ap \in APs} \sum_{i \in L} (\mathbf{c}_{i}-\mathbf{c}_{i}^{ap})\tp \int_{t_{s,i,ap}}^{t_{e,i,ap}}\mathbf{t} \mathbf{t}\tp dt \ (\mathbf{c}_{i}-\mathbf{c}_{i}^{ap}) \\
=&\sum_{ap \in APs} (\mathbf{c}-\mathbf{c}^{ap})\tp \mathbf{Q}_{c,ap} (\mathbf{c}-\mathbf{c}^{ap}),
%=&(\mathbf{c}-\mathbf{c}^{AP})\tp \mathbf{Q}_{c} (\mathbf{c}-\mathbf{c}^{AP}),
\end{aligned}
\end{equation}
where $p_{ap}(t)$ is the constant positioin of one attracting point $ap$ out of the set $APs$, and $(t_{e,ap}$ - $t_{s,ap}) \subseteq [0, T]$ is the corresponding time period of the optimized trajectory that is affected by the dragging force.
$L$ is the set of specific piece indices that is drawn by an attracting point with $(t_{e,i,ap}$ - $t_{s,i,ap}) \subseteq [0, T_i]$ the involved time period in the $i^{th}$ piece.

The overall objective is formed as a weighted sum of the three terms, and the optimization problem is formulated in the following:
\begin{equation}
\begin{aligned}
\min J =&\ \lambda_{s} J_{s}+\lambda_{r} J_{r}+\lambda_{c} J_{c} \\[1ex]
=& [\mathbf{c}\tp(\lambda_{s}\mathbf{Q}_s+\lambda_{h}\mathbf{Q}_h+\lambda_{c} \sum_{ap \in APs} \mathbf{Q}_{c,ap})\mathbf{c} \\
& -2 \mathbf{c}\tp(\lambda_{r} \mathbf{Q}_r\mathbf{c}^* +\lambda_{c} \sum_{ap \in APs} \mathbf{Q}_{c,ap}\mathbf{c}^{ap}) \\
& +\lambda_{r}(\mathbf{c}^*)\tp \mathbf{Q}_r\mathbf{c}^* +\lambda_{c} \sum_{ap \in APs} (\mathbf{c}^{ap})\tp \mathbf{Q}_{c,ap}\mathbf{c}^{ap}] \\[1ex]
s.t. \quad \mathbf{A}&\mathbf{c}=\mathbf{d},\\
\forall t& \in [0, \tau],\
\mathbf{x}(t) \in \mathcal{X}^{free}, \
\mathbf{u}(t) \in \mathcal{U}^{free},
\end{aligned}
\end{equation}
where $\lambda_{s}$, $\lambda_{r}$, and $\lambda_{c}$ are the respective weights, $c$ the decision variable vector, and $\mathbf{A}\mathbf{c}=\mathbf{d}$ the boundary derivative constraints for each pieces.

In this way, we seek a smoother trajectory that is close to the original one for the collision-free parts while deforms to neighbor collision-free areas for the collided parts.

\subsection{Iterative Optimization Process}

\begin{figure}[t]
\centering
\begin{subfigure}{0.493\linewidth}
	\includegraphics[width=1\linewidth]{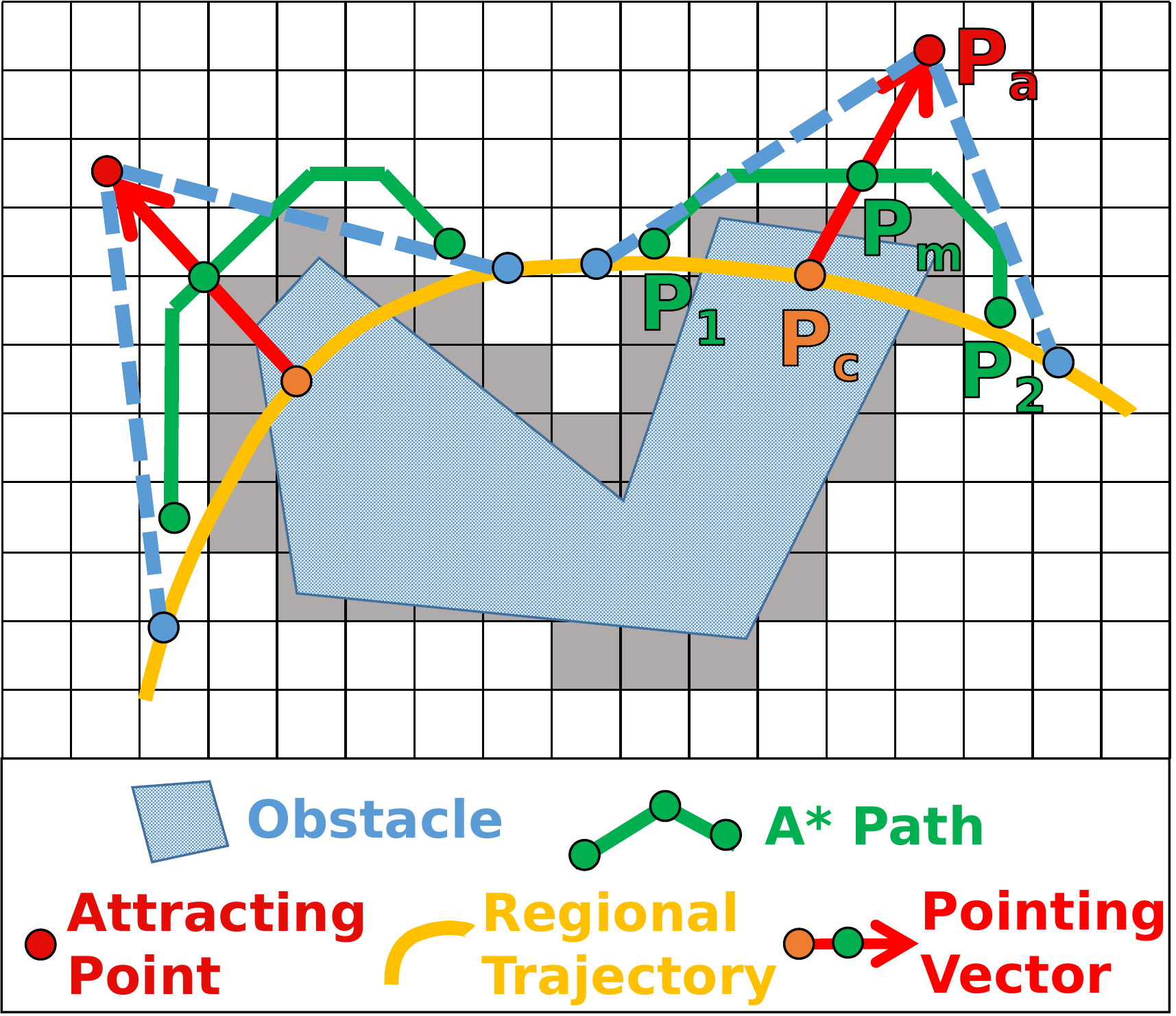}	
	\caption{Regional Optimizer Cases}
	\label{subfig:regional_ap}
\end{subfigure}
\begin{subfigure}{0.493\linewidth}
	\includegraphics[width=1\linewidth]{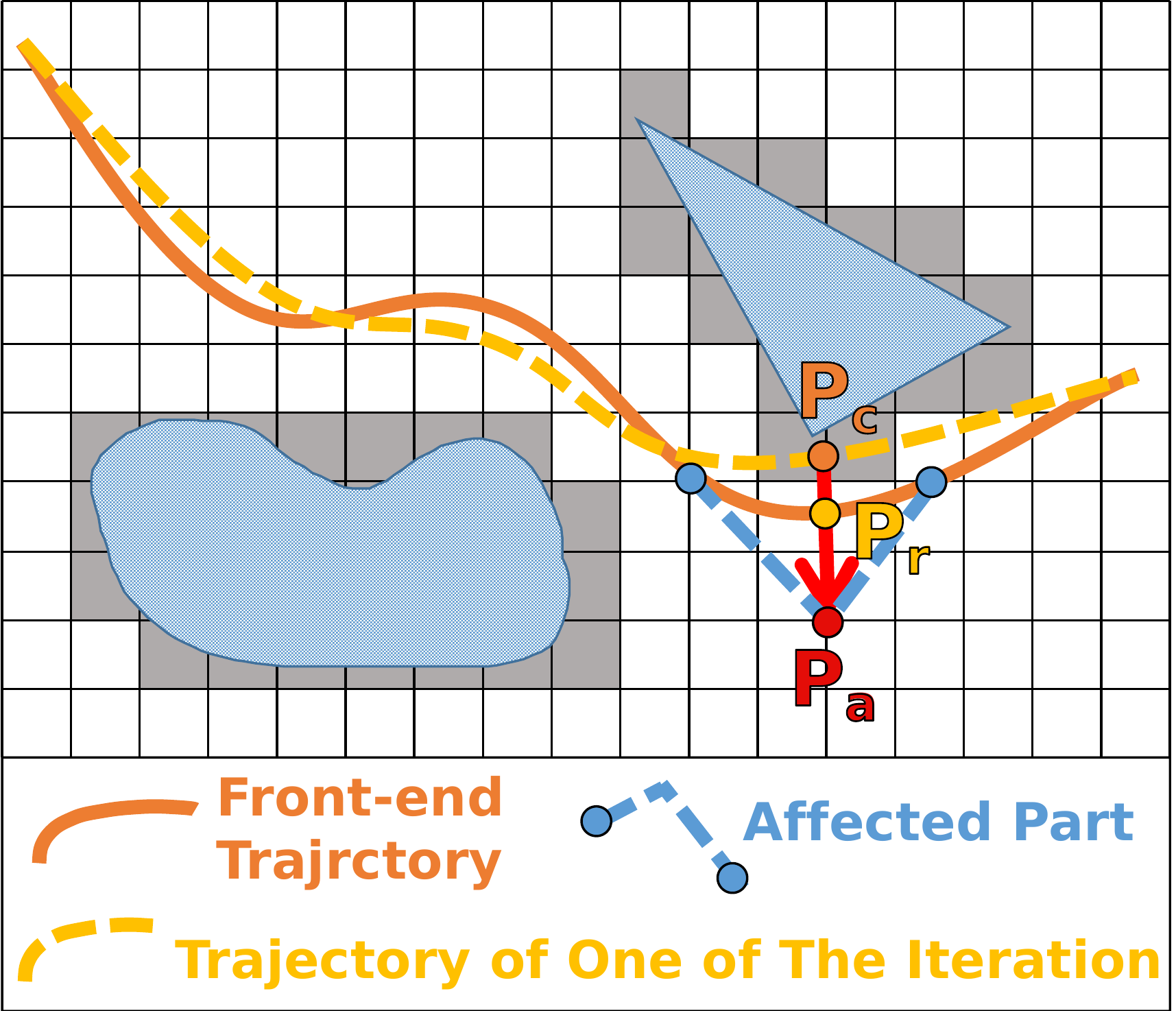}
	\caption{Back-end Cases}
	\label{subfig:backend_ap}
\end{subfigure}
\captionsetup{font={small}}
\caption{Illustration of selecting attracting points in 2D, the method applies for 3D environments as well.
(a) For the regional optimization cases, when the regional trajectory (yellow curve) collides with obstacles, we denote the start and endpoint of collision as $P_1$ and $P_2$, the point in the middle of the collision part as $P_c$. Then we search a free path (green lines) from $P_1$ to $P_2$ and denote the middle point in the path as $P_m$. The attracting point $P_a$ is chosen in some distance in the extending line of the pointing vector (red arrow) from $P_c$ to $P_m$.
(b) For the back-end optimization cases, when the temporal result (yellow dashed curve) of an iteration collides, we find a correspondent point $P_r$ of the middle collision point $P_c$ in the front-end trajectory (orange curve) and draw a pointing vector from $P_r$ to $P_c$. The attracting point is chosen in its extending direction.}
\label{fig:att_points_selection}
\vspace{-0.7cm}
\end{figure}

As we can see, the overall objective $J$ is quadratic, and the coefficient matrix of the quadratic term is always positive definite if the weights and time durations are non-negative.
Following \cite{bry2015aggressive, wang2020generating} and utilizing the PD property, an unconstrained formulation of QP incorporating the boundary derivative constraints can be derived. Ignoring the state and control constraints, the optimal solution can be obtained in closed-form with given boundary conditions and time allocation, which is the process of \textbf{ClosedFormSolve}() in Alg.~\ref{alg:RegionalOptimize}.
We then check for state and control saturation and obstacle feasibility of the unconstrained solution.
If it collides with new obstacles, we reformulate the \textit{Collision Cost} by incrementally adding new selected attracting points to provide more accurate dragging forces and information of the local surrounding environment.
If the state and control violate saturations, we increase the time duration of the whole trajectory.
This process runs iteratively until a feasible trajectory is found or maximum iteration time is reached.

\begin{figure}[t]
\centering
\includegraphics[width=0.9\linewidth]{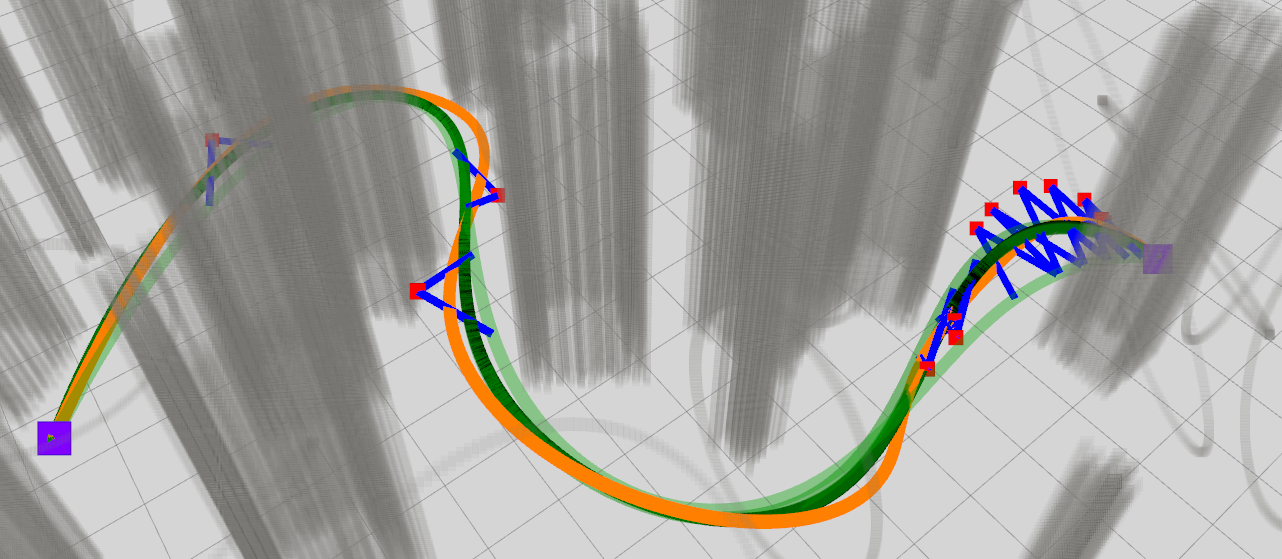}
\captionsetup{font={small}}
\caption{Visualization of the iteration process of back-end optimization. Obstacles are set transparent to provide better views. The orange curve is the front-end trajectory. The red dots represent attracting points, with blue lines indicating the affected parts. As iterations proceed, the previous attracting points preserve while more are added, and the optimized trajectory is shown as green curves colored from light to dark.}
\label{fig:back_end_opt_process}
\vspace{-0.8cm}
\end{figure}

The selection of the attracting points is important since they guide the deformation of the optimized trajectory. For each collision part of the checked trajectory, we use local search methods like A* to find an obstacle-free path around it, and the attracting point for this collision part is chosen at some distance of the vector, which points from the middle collision position to the middle position of the A* path. This attracting point will only affect a nearby part of the trajectory around the collision part. Fig.~\ref{subfig:regional_ap} depicts this procedure. Since we only do regional optimization for relatively short local trajectories, the search area is usually very restricted, and grid search finishes within microseconds.
In~\cite{Zhou2021EGO}, the authors use a similar local search strategy to acquire rough distance gradient information, which can harm the convergence if possibly inaccurate. Our formulation, however, avoids this by deducing a closed-form solution and requires no gradient.
The objective in our problem changes in each iteration instead.
A visualization of the iterative optimization process is shown in Fig.~\ref{fig:back_end_opt_process}.

%To further specify the calling of this optimizer in the global sampling process and avoid wasting computation on non-promising samples, we set conditions that the optimizer is only triggered if the trajectory length is less than some predefined metric and only if a small part of the trajectory is in collision. How to generate promising samples in the first place is another way to improve the planning efficiency and is left for future work.

\subsection{Trajectory Refinement}
From the kinodynamic front-end that globally seeks an asymptotically optimal trajectory, we have an initial trajectory that settles in an appropriate homotopy class and satisfies all the constraints with the time duration reasonably allocated for each segment.
However, it may not be smooth enough because of insufficient sampling within a limited time budget. As many other works have validated, a lightweight refinement can greatly improve the result.

In our case, we use the same optimization framework developed for the regional optimizer.
One of the differences is that the search for a collision-free path is no longer needed when selecting attracting points.
This is because the initial trajectory is now collision-free already.
When the resulting trajectory of one iteration collides with obstacles, a point in the extending line of the vector which points from the middle collision point to the corresponding point in the initial collision-free trajectory is chosen as the attracting point for this collision part, as illustrated in Fig.~\ref{subfig:backend_ap}.
By addressing the corresponding point in the initial trajectory, we mean the position at the same time instance when the collision occurs.
This correspondence is based on the added \textit{Resemblance Cost} term and the fact that we use the front-end time allocation to initialize the back-end. This engenders similarities between the initial trajectory and the optimized trajectory.
By fully exploiting the kinodynamic front-end assets and taking obstacle clearance into account while still retaining a closed-form solution for each iteration, this refinement is extremely fast and remarkably improves the success rate compared to our previous work~\cite{Hongkai2021tgk}.

%Unlike in the regional optimization, it is a global optimization in the back-end where the trajectories are much longer.

\begin{figure}[t]
\centering
\begin{subfigure}{0.95\linewidth}
	\includegraphics[width=1\linewidth]{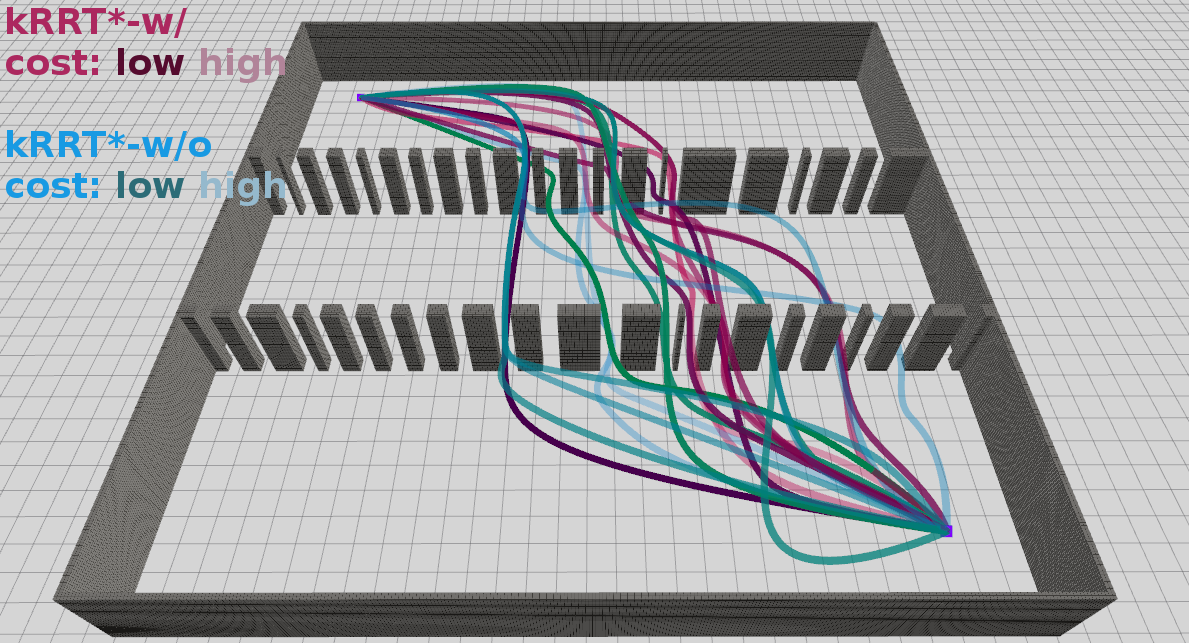}
	\caption{Intermediate trajectories in $10s$ planning time of one trial. The blue ones are results of kRRT* integrating RO, and the red ones are results for the other. Trajectories obtained earlier are drawn in a lighter color. Different homotopy classes are explored and the solution converges to the optimum as planning proceeds.}
	\label{fig:w_wo}
	\vspace{0.3cm}
\end{subfigure}
\begin{subfigure}{0.9\linewidth}
	\includegraphics[width=1\linewidth]{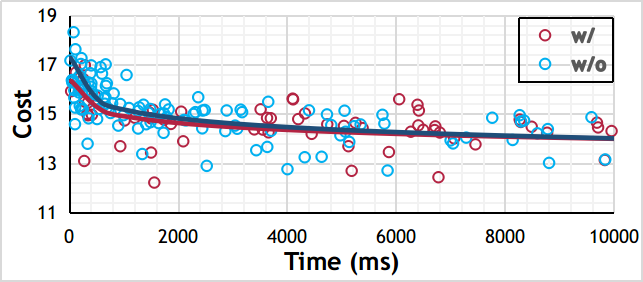}
	\caption{kRRT* w/ RO has a higher convergence rate.}	
	\label{fig:w_wo_data}
\end{subfigure}
\captionsetup{font={small}}
\caption{Comparison of kRRT* w/ and w/o RO integrated.}
\vspace{-1.0cm}
\end{figure}

\section{Benchmark and Experiment}
\label{sec:experiment}
\subsection{Experiment Settings}
We set the state and control limitations of our multirotor as $7m/s$ for velocity, $5m/s^2$ for acceleration, and $15m/s^3$ for jerk. The weight of time $\rho$ is set $100$.
For collision checking, we use occupancy grid maps of $0.1m$ resolution, and the obstacles are inflated by $0.3m$, which is the radius of our multirotor.
All the benchmark computations are done with a 3.4GHz Intel i7-6700 processor.

\begin{figure*}[t]
\centering
\begin{subfigure}{0.95\linewidth}
	\includegraphics[width=1\linewidth]{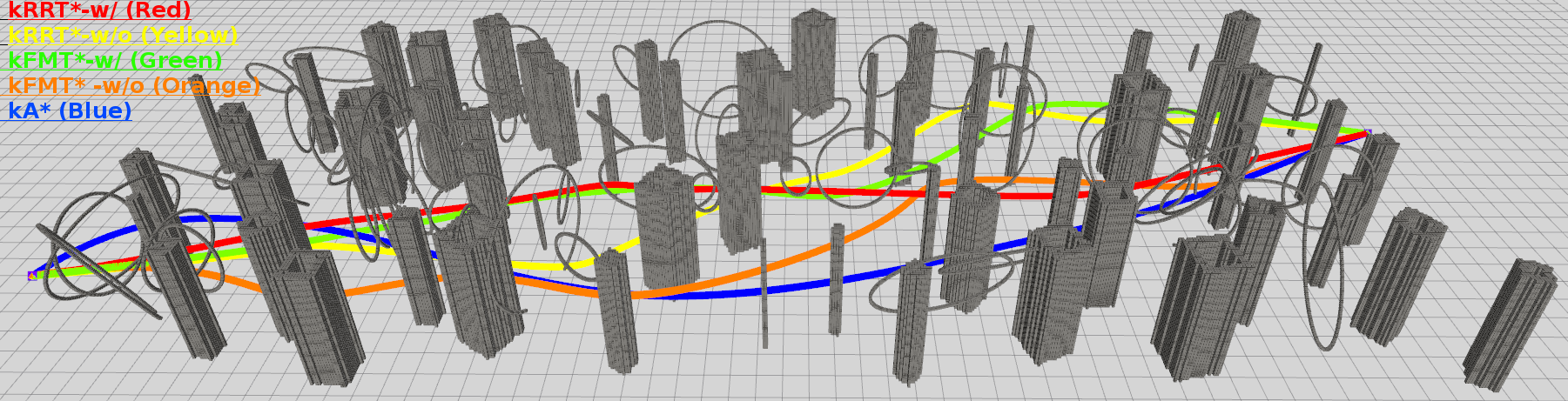}
	\caption{$2.5D$ Forest-like environments.}
	\label{fig:forest}
\end{subfigure}
\begin{subfigure}{0.95\linewidth}
	\includegraphics[width=1\linewidth]{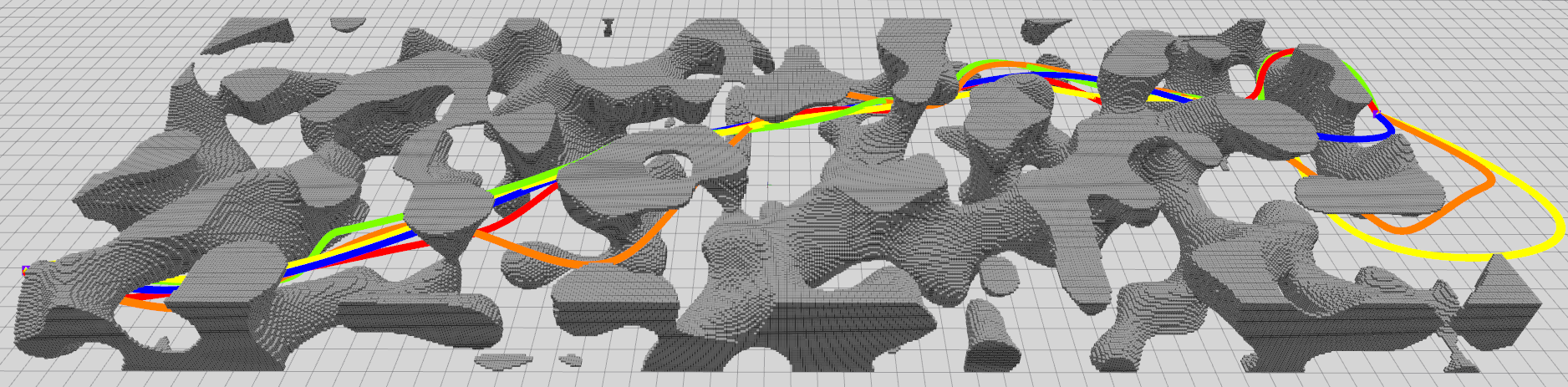}
	\caption{$3D$ Cave-like environments with complex homotopy classes. }	
	\label{fig:cave}
\end{subfigure}
\begin{subfigure}{0.95\linewidth}
	\includegraphics[width=1\linewidth]{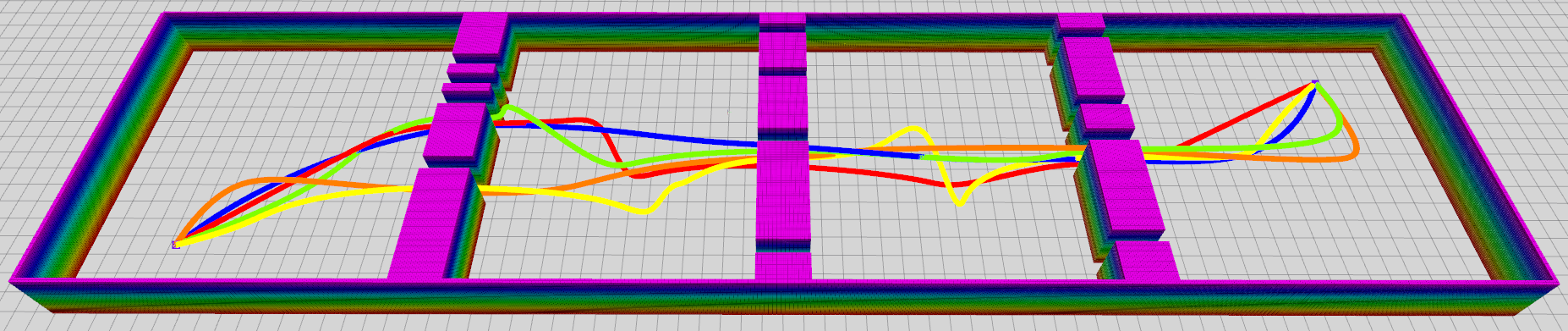}
	\caption{$2.5D$ Corridor environments with only narrow passages to travel through.}
	\label{fig:wall}
\end{subfigure}
\captionsetup{font={small}}
\caption{Front-end trajectories in different environments. The color indications are red for kRRT*-w/ (proposed), yellow for kRRT*-w/o, green for kFMT*-w/, orange for kFMT*-w/o, and blue for kA*.}
\label{fig:frontend_compare}
\end{figure*}

\subsection{Global Sampling w/ and w/o Regional Optimization}
To validate the effectiveness of the integration, the kRRT* methods w/ and w/o the integration of the regional optimization are compared.
The test environment is shown in Fig.~\ref{fig:w_wo}, which is a $30m \times 30m \times 3m$ box separated by two walls, and each wall is split by 20 gaps to create many different homotopy classes. The gap width is $0.7m$, which just fits the multirotor to travel through after inflation.
We set a $10s$ time budget for each of the $100$ trials for both methods and record the solution cost as the planning proceeds. The convergence trend is plotted in Fig.~\ref{fig:w_wo_data}. As shown, the kRRT* method integrated with the proposed regional optimization converges to the optimum quicker than the other. Besides, the first solution obtained has a lower cost. The results validate the effectiveness of exploiting local domain information in the global sampling process. One instance of the intermediate trajectories obtained during planning can be seen in Fig.~\ref{fig:w_wo}.

\subsection{Front-end Comparison}

\begin{table*}[t]
\centering
\captionsetup{font={small}}
\caption{Front-end comparison results.}
\label{tab:front_end_data}
\setlength{\tabcolsep}{1.7mm}{
\begin{tabular}{cccccccccccccccccc}
                                                            & \multicolumn{2}{c}{\textit{\textbf{Col.}}}                              & \textit{\textbf{1}}         & \textit{\textbf{2}}          & \textit{\textbf{3}}          & \textit{\textbf{4}}        & \textit{\textbf{5}}        & \textit{\textbf{6}}        & \textit{\textbf{7}}         & \textit{\textbf{8}}         & \textit{\textbf{9}}         & \textit{\textbf{10}}       & \textit{\textbf{11}}       & \textit{\textbf{12}}       & \textit{\textbf{13}}       & \textit{\textbf{14}}       & \textit{\textbf{15}}       \\ \cline{2-18}
\multicolumn{1}{c|}{\multirow{2}{*}{\textit{\textbf{Row}}}} & \multicolumn{2}{c|}{\multirow{2}{*}{Method}}                            & \multicolumn{3}{c|}{\begin{tabular}[c]{@{}c@{}}Planning Time\\ (ms)\end{tabular}}         & \multicolumn{3}{c|}{\begin{tabular}[c]{@{}c@{}}Trajectory\\ Length (m)\end{tabular}} & \multicolumn{3}{c|}{\begin{tabular}[c]{@{}c@{}}Trajectory \\ Duration (s)\end{tabular}} & \multicolumn{3}{c|}{Cost / 100}                                                      & \multicolumn{3}{c|}{\begin{tabular}[c]{@{}c@{}}Success Rate\\  (\%)\end{tabular}}    \\ \cline{4-18}
\multicolumn{1}{c|}{}                                       & \multicolumn{2}{c|}{}                                                   & \multicolumn{1}{c|}{F.}     & \multicolumn{1}{c|}{C.}      & \multicolumn{1}{c|}{W.}      & \multicolumn{1}{c|}{F.}    & \multicolumn{1}{c|}{C.}    & \multicolumn{1}{c|}{W.}    & \multicolumn{1}{c|}{F.}     & \multicolumn{1}{c|}{C.}     & \multicolumn{1}{c|}{W.}     & \multicolumn{1}{c|}{F.}    & \multicolumn{1}{c|}{C.}    & \multicolumn{1}{c|}{W.}    & \multicolumn{1}{c|}{F.}    & \multicolumn{1}{c|}{C.}    & \multicolumn{1}{c|}{W.}    \\ \cline{2-18}
\multicolumn{1}{c|}{\textit{\textbf{1}}}                    & \multicolumn{1}{c|}{\multirow{2}{*}{kRRT*}} & \multicolumn{1}{c|}{-w/}  & \multicolumn{1}{c|}{46.5}   & \multicolumn{1}{c|}{1444.5}  & \multicolumn{1}{c|}{2075.6}  & \multicolumn{1}{c|}{66.9}  & \multicolumn{1}{c|}{66.8}  & \multicolumn{1}{c|}{64.9}  & \multicolumn{1}{c|}{19.4}   & \multicolumn{1}{c|}{19.5}   & \multicolumn{1}{c|}{18.9}   & \multicolumn{1}{c|}{24.31} & \multicolumn{1}{c|}{24.88} & \multicolumn{1}{c|}{23.45} & \multicolumn{1}{c|}{100.0} & \multicolumn{1}{c|}{84.33} & \multicolumn{1}{c|}{96.67} \\ \cline{3-18}
\multicolumn{1}{c|}{\textit{\textbf{2}}}                    & \multicolumn{1}{c|}{}                       & \multicolumn{1}{c|}{-w/o} & \multicolumn{1}{c|}{60.6}   & \multicolumn{1}{c|}{1678.6}  & \multicolumn{1}{c|}{2548.7}  & \multicolumn{1}{c|}{67.9}  & \multicolumn{1}{c|}{67.5}  & \multicolumn{1}{c|}{65.3}  & \multicolumn{1}{c|}{20.0}   & \multicolumn{1}{c|}{19.8}   & \multicolumn{1}{c|}{18.4}   & \multicolumn{1}{c|}{24.78} & \multicolumn{1}{c|}{24.70} & \multicolumn{1}{c|}{22.76} & \multicolumn{1}{c|}{99.33} & \multicolumn{1}{c|}{75.67} & \multicolumn{1}{c|}{74.33} \\ \cline{2-18}
\multicolumn{1}{c|}{\textit{\textbf{3}}}                    & \multicolumn{1}{c|}{\multirow{2}{*}{kFMT*}} & \multicolumn{1}{c|}{-w/}  & \multicolumn{1}{c|}{1610.4} & \multicolumn{1}{c|}{1127.9}  & \multicolumn{1}{c|}{1888.6}  & \multicolumn{1}{c|}{64.8}  & \multicolumn{1}{c|}{67.3}  & \multicolumn{1}{c|}{65.0}  & \multicolumn{1}{c|}{17.8}   & \multicolumn{1}{c|}{20.5}   & \multicolumn{1}{c|}{19.1}   & \multicolumn{1}{c|}{22.28} & \multicolumn{1}{c|}{26.83} & \multicolumn{1}{c|}{24.82} & \multicolumn{1}{c|}{81.67} & \multicolumn{1}{c|}{96.33} & \multicolumn{1}{c|}{44.00} \\ \cline{3-18}
\multicolumn{1}{c|}{\textit{\textbf{4}}}                    & \multicolumn{1}{c|}{}                       & \multicolumn{1}{c|}{-w/o} & \multicolumn{1}{c|}{1524.9} & \multicolumn{1}{c|}{1127.0}  & \multicolumn{1}{c|}{1885.1}  & \multicolumn{1}{c|}{65.0}  & \multicolumn{1}{c|}{67.5}  & \multicolumn{1}{c|}{65.2}  & \multicolumn{1}{c|}{17.6}   & \multicolumn{1}{c|}{20.3}   & \multicolumn{1}{c|}{19.1}   & \multicolumn{1}{c|}{21.52} & \multicolumn{1}{c|}{25.40} & \multicolumn{1}{c|}{23.80} & \multicolumn{1}{c|}{89.00} & \multicolumn{1}{c|}{94.67} & \multicolumn{1}{c|}{32.67} \\ \cline{2-18}
\multicolumn{1}{c|}{\textit{\textbf{5}}}                    & \multicolumn{2}{c|}{kA*}                                                & \multicolumn{1}{c|}{6979.9} & \multicolumn{1}{c|}{89802.0} & \multicolumn{1}{c|}{37624.0} & \multicolumn{1}{c|}{64.1}  & \multicolumn{1}{c|}{71.6}  & \multicolumn{1}{c|}{64.1}  & \multicolumn{1}{c|}{13.1}   & \multicolumn{1}{c|}{16.1}   & \multicolumn{1}{c|}{13.4}   & \multicolumn{1}{c|}{14.16} & \multicolumn{1}{c|}{18.71} & \multicolumn{1}{c|}{14.58} & \multicolumn{1}{c|}{100.0} & \multicolumn{1}{c|}{81.00} & \multicolumn{1}{c|}{98.33} \\ \cline{2-18}
\end{tabular}
}
\end{table*}

\begin{table*}[t]
\centering
\captionsetup{font={small}}
\caption{Back-end comparison results.}
\label{tab:back-end}
\begin{tabular}{|c|c|c|c|c|c|c|c|c|c|c|c|c|c|c|c|}
\hline
\multirow{2}{*}{Method} & \multicolumn{3}{c|}{\begin{tabular}[c]{@{}c@{}}Planning Time\\ (ms)\end{tabular}} & \multicolumn{3}{c|}{\begin{tabular}[c]{@{}c@{}}Iteration \\ Times (1)\end{tabular}} & \multicolumn{3}{c|}{\begin{tabular}[c]{@{}c@{}}Trajectory\\ Length (m)\end{tabular}} & \multicolumn{3}{c|}{\begin{tabular}[c]{@{}c@{}}Jerk Integration\\ (m2 / s5)\end{tabular}} & \multicolumn{3}{c|}{\begin{tabular}[c]{@{}c@{}}Success Rate\\  (\%)\end{tabular}} \\ \cline{2-16}
                        & F.                                      & C.                 & W.                 & F.                         & C.                         & W.                        & F.                         & C.                         & W.                         & F.                           & C.                           & W.                          & F.                        & C.                        & W.                        \\ \hline
Proposed                & \multicolumn{1}{l|}{\textbf{0.27}}      & \textbf{0.84}      & 0.79               & \textbf{2.0}               & 5.1                        & \textbf{7.9}              & \textbf{64.1}              & \textbf{71.7}              & \textbf{66.8}              & \textbf{132.1}               & \textbf{187.9}               & \textbf{171.5}              & \textbf{97.33}            & \textbf{94.00}            & 89.33                     \\ \hline
Base 1                 & 0.47                                    & 1.16               & \textbf{0.69}      & 8.1                        & 10.9                       & 10.9                      & 64.3                       & 72.9                       & 67.4                       & 215.3                        & 584.8                        & 437.2                       & 94.67                     & 76.67                     & 55.33                     \\ \hline
Base 2                 & 0.58                                    & 1.71               & 1.04               & 2.9                        & \textbf{3.2}               & 3.5                       & 64.9                       & 72.9                       & 67.8                       & 171.5                        & 331.8                        & 237.4                       & 97.00                     & 88.33                     & \textbf{95.67}            \\ \hline
Base 3                 & 12.1                                    & 20.1               & 13.0               & 8.2                        & 8.1                        & 8.9                       & 65.0                       & 72.4                       & 66.9                       & 573.7                        & 721.6                        & 930.4                       & 62.33                     & 92.67                     & 21.33                     \\ \hline
\end{tabular}
%}
\end{table*}

We compare our front-end (kRRT*-w/) with four other kinodynamic planning methods with the same dynamics and the same objective. They are 1) kRRT* without integrating RO (kRRT*-w/o)~\cite{webb2013kinodynamic, Hongkai2021tgk}, 2) kinodynamic A* (kA*)~\cite{liu2017iros} that uses jerk as control input, 3) kinodynamic fast marching tree* integrating the proposed RO (kFMT*-w/)~\cite{Janson2015FMT, Schmerling2015drift}, and 4) kFMT* without integrating RO (kFMT*-w/o).
For kA*, we adopt $5$ equally discretized inputs and integrate the dynamics for a time period of $0.5s$.
For kFMT*-w/, the regional optimization is performed in each of its lazy dynamic programming step.~\cite{Janson2015FMT}.
The same sampling strategy~\cite{Hongkai2021tgk} is adopted for all kRRT* and kFMT* style methods.
Each method runs for $300$ trials with different starts and goals in three kinds of environments, which are 1) a $2.5D$ forest-like one (F.) with randomly placed obstacles, 2) a $3D$ cave-like one (C.) with more complex homotopy classes, and 3) a $2.5D$ corridor (W.) blocked by $3$ thick walls with random narrow passages in it. The narrow passages and the wall thickness further reduce the possibility of directly connecting two samples without regional optimization.
The environments and the result trajectories of one trial are shown in Fig.~\ref{fig:frontend_compare}.

The effectiveness of integrating the proposed RO can be validated by comparing \textit{Row 1} with \textit{Row 2}, \textit{Row 3} with \textit{Row 4}, Tab.\ref{tab:front_end_data}, which shows that kRRT*-w/ presents much better statistics than kRRT*-w/o among all the test cases, especially in planning time and success rate, and that kFMT*-w/ achieves a higher success rate with comparable planning time than kFMT*-w/o in the cave and corridor cases where more narrow passages present.
However, in the forest-like environment where there are much more open areas, kFMT*-w/ shows slight inferiority to kFMT*-w/o. We think it can be caused by unnecessary optimizations on nonpromising samples, suggesting that there is space for improvement in the integration manner.

To study the effectiveness of integrating RO with different sampling-based methods, we compare \textit{Row 1} with \textit{Row 3}, \textit{Row 2} with \textit{Row 4}, Tab.\ref{tab:front_end_data}.
kRRT* samples incrementally, and its performance is related to the sampling sequence, while kFMT* samples in batch and then performs the search.
In the forest-like environment, the proposed method kRRT*-w/ achieves the highest success rate with the shortest planning time.
In the cave-like environment, kRRT* requires more planning time and has less success rate than kFMT*. We think it is because in long-range global planning, kRRT*'s incremental sampling produces more wasted samples if they are far from the tree in the solution space, and the kRRT* tree growing can be stuck in the middle if there are many narrow spaces.
We notice that in the corridor environment, the success rate of kRRT* prevails by double (\textit{Col. 15}, Tab.\ref{tab:front_end_data}). This is caused by the kFMT*'s limited and unchanged samples drawn in batch, which harms the tree growing when there are only narrow gaps to plan through. Integrating the proposed RO improves it.

For the comparison with the search-based method (\textit{Row 5}, Tab.\ref{tab:front_end_data}), we notice that although kA* returns solutions with the lowest cost and the shortest trajectory duration, it takes untractable planning time (\textit{Col. 1, 2, 3}, Tab.\ref{tab:front_end_data}) which is hundreds of times than the proposed method. Being asymptotically optimal, the proposed method returns a feasible solution quickly within milliseconds, and the solution improves with an extra time budget and thus is applicable for real-time usages.

\subsection{Back-end Comparison}

To evaluate the manners of investigating the front-end results, benchmarks are conducted with $3$ other back-end methods. They are 1) our previous work (Base 1)~\cite{Hongkai2021tgk} that formulates a \textit{Homotopy Cost} to force the optimized trajectory to stay in nearby free spaces but considers no surrounding obstacle information, 2) Liu's method (Base 2)~\cite{bry2015aggressive, liu2017iros} that only utilizes fixed intermediate waypoints and time allocation from the front-end, and 3) Zhou's method (Base 3)~\cite{boyu2019ral} that samples control points from the front-end trajectory and reparameterizes the result as a B-spline, and requires a distance field computed a prior (about $150ms$ in the test cases) to push the trajectory into collision-free areas.
By definition, Base 1, Base 2, and the proposed method all have a closed-form solution in each iteration, whereas Base 3 formulates a soft-constrained nonlinear optimization problem that uses NLopt~\cite{nlopt} to solve.
The test environments are the same as those used in the previous section but with different random seeds.
We run $300$ trials for each test case, and the same front-end result of kRRT*-w/ is used for each trial.

As shown in Tab.~\ref{tab:back-end}, the proposed method dominates in almost all the criteria except that in the corridor environments, the success rate is a bit lower than Base 2's. This is because in extremely confined environments like long narrow passages with a width that just fits the multirotor, adding fixed waypoints from the front-end trajectory in each iteration like Base 2 dose can provide stronger constraints to force the optimized trajectory to stay in the free narrow passages.
Compared to others, the overall high performance of the proposed method can be explained by several ideas that 1) the incorporating of nearby collisions as \textit{Collision Cost} provides richer information of the environments, and deforms the parts of the trajectory where it is originally near obstacles into obstacle-free areas, resulting in much higher success rate compared to Base 1, and 2) no requirement for any fixed waypoints provides more freedom of deformation, obtaining smoother trajectories with much lower jerk integration.
Base 3, however, requires much more time for the nonlinear programming to converge and fails a lot if no accurate distance field is available, which is usually the case in cluttered and confined environments.
Under circumstances where a good front-end trajectory can be acquired, the proposed back-end methods can act as a great complement, providing much smoother trajectories with little effort.

\section{Conclusion \& Future Work}
We present a kinodynamic planning method and framework that meet real-time requirements for multirotor flight. The front-end globally reasons for an asymptotically optimal trajectory by the modified kRRT* integrating an efficient and effective regional optimizer, which better handles difficult-to-sample cases. The proposed regional optimizer is reformed as the back-end refiner, which better exploits the front-end result and greatly improves it with negligible computation effort. Benchmark results show the superiority of the proposed methods against other state-of-the-art multirotor kinodynamic planning methods. This work again validates the effectiveness of integrating regional optimization into the global sampling process but still leaves space for improvement. We plan to investigate further the manners of integration on other batch sampling-based planners.
We are also interested in incorporating attitude constraints to realize fully $\mathit{SE}(3)$ planning.

\bibliography{main}

\begin{thebibliography}{10}
\providecommand{\url}[1]{#1}
\csname url@rmstyle\endcsname
\providecommand{\newblock}{\relax}
\providecommand{\bibinfo}[2]{#2}
\providecommand\BIBentrySTDinterwordspacing{\spaceskip=0pt\relax}
\providecommand\BIBentryALTinterwordstretchfactor{4}
\providecommand\BIBentryALTinterwordspacing{\spaceskip=\fontdimen2\font plus
\BIBentryALTinterwordstretchfactor\fontdimen3\font minus
  \fontdimen4\font\relax}
\providecommand\BIBforeignlanguage[2]{{%
\expandafter\ifx\csname l@#1\endcsname\relax
\typeout{** WARNING: IEEEtran.bst: No hyphenation pattern has been}%
\typeout{** loaded for the language `#1'. Using the pattern for}%
\typeout{** the default language instead.}%
\else
\language=\csname l@#1\endcsname
\fi
#2}}

\bibitem{mueller2015computationally}
M.~W. Mueller, M.~Hehn, and R.~D'Andrea, ``A computationally efficient motion
  primitive for quadrocopter trajectory generation,'' \emph{IEEE Transactions
  on Robotics}, vol.~31, no.~6, pp. 1294--1310, 2015.

\bibitem{liu2018ral}
S.~{Liu}, K.~{Mohta}, N.~{Atanasov}, and V.~{Kumar}, ``Search-based motion
  planning for aggressive flight in se(3),'' \emph{IEEE Robotics and Automation
  Letters}, vol.~3, no.~3, pp. 2439--2446, 2018.

\bibitem{kaufmann2020RSS}
E.~Kaufmann, A.~Loquercio, R.~Ranftl, M.~M{\"u}ller, V.~Koltun, and
  D.~Scaramuzza, ``Deep drone acrobatics,'' \emph{RSS: Robotics, Science, and
  Systems}, 2020.

\bibitem{Donald1993kinodynamic}
\BIBentryALTinterwordspacing
B.~Donald, P.~Xavier, J.~Canny, and J.~Reif, ``Kinodynamic motion planning,''
  \emph{J. ACM}, vol.~40, no.~5, p. 1048–1066, Nov. 1993. [Online].
  Available: \url{https://doi.org/10.1145/174147.174150}
\BIBentrySTDinterwordspacing

\bibitem{liu2017iros}
S.~Liu, N.~Atanasov, K.~Mohta, and V.~Kumar, ``Search-based motion planning for
  quadrotors using linear quadratic minimum time control,'' in \emph{Proc. of
  the {IEEE/RSJ} Intl. Conf. on Intell. Robots and Syst.}, Sept 2017, pp.
  2872--2879.

\bibitem{likhachev2009planning}
M.~Likhachev and D.~Ferguson, ``Planning long dynamically feasible maneuvers
  for autonomous vehicles,'' \emph{The International Journal of Robotics
  Research}, vol.~28, no.~8, pp. 933--945, 2009.

\bibitem{Schmerling2015drift}
E.~{Schmerling}, L.~{Janson}, and M.~{Pavone}, ``Optimal sampling-based motion
  planning under differential constraints: The drift case with linear affine
  dynamics,'' in \emph{2015 54th IEEE Conference on Decision and Control
  (CDC)}, 2015, pp. 2574--2581.

\bibitem{Schmerling2015driftless}
------, ``Optimal sampling-based motion planning under differential
  constraints: The driftless case,'' in \emph{2015 IEEE International
  Conference on Robotics and Automation (ICRA)}, 2015, pp. 2368--2375.

\bibitem{webb2013kinodynamic}
D.~J. {Webb} and J.~{van den Berg}, ``Kinodynamic rrt*: Asymptotically optimal
  motion planning for robots with linear dynamics,'' in \emph{Proc. of the
  {IEEE} Intl. Conf. on Robot. and Autom.}, May 2013, pp. 5054--5061.

\bibitem{Hongkai2021tgk}
H.~{Ye}, X.~{Zhou}, Z.~{Wang}, C.~{Xu}, J.~{Chu}, and F.~{Gao}, ``Tgk-planner:
  An efficient topology guided kinodynamic planner for autonomous quadrotors,''
  \emph{IEEE Robotics and Automation Letters}, vol.~6, no.~2, pp. 494--501,
  2021.

\bibitem{Choudhury2016Regionally}
S.~{Choudhury}, J.~D. {Gammell}, T.~D. {Barfoot}, S.~S. {Srinivasa}, and
  S.~{Scherer}, ``Regionally accelerated batch informed trees (rabit*): A
  framework to integrate local information into optimal path planning,'' in
  \emph{2016 IEEE International Conference on Robotics and Automation (ICRA)},
  2016, pp. 4207--4214.

\bibitem{Kim2018Dancing}
D.~{Kim}, Y.~{Kwon}, and S.~{Yoon}, ``Dancing prm*: Simultaneous planning of
  sampling and optimization with configuration free space approximation,'' in
  \emph{2018 IEEE International Conference on Robotics and Automation (ICRA)},
  2018, pp. 7071--7078.

\bibitem{Kim2019Volumetric}
D.~{Kim}, M.~{Kang}, and S.~{Yoon}, ``Volumetric tree*: Adaptive sparse graph
  for effective exploration of homotopy classes,'' in \emph{2019 IEEE/RSJ
  International Conference on Intelligent Robots and Systems (IROS)}, 2019, pp.
  1496--1503.

\bibitem{dolgov2010path}
D.~Dolgov, S.~Thrun, M.~Montemerlo, and J.~Diebel, ``Path planning for
  autonomous vehicles in unknown semi-structured environments,'' \emph{The
  International Journal of Robotics Research}, vol.~29, no.~5, pp. 485--501,
  2010.

\bibitem{boyu2019ral}
B.~Zhou, F.~Gao, L.~Wang, C.~Liu, and S.~Shen, ``Robust and efficient quadrotor
  trajectory generation for fast autonomous flight,'' \emph{IEEE Robotics and
  Automation Letters}, vol.~4, no.~4, pp. 3529--3536, 2019.

\bibitem{Gao2018Online}
F.~{Gao}, W.~{Wu}, Y.~{Lin}, and S.~{Shen}, ``Online safe trajectory generation
  for quadrotors using fast marching method and bernstein basis polynomial,''
  in \emph{2018 IEEE International Conference on Robotics and Automation
  (ICRA)}, 2018, pp. 344--351.

\bibitem{bry2015aggressive}
A.~Bry, C.~Richter, A.~Bachrach, and N.~Roy, ``Aggressive flight of fixed-wing
  and quadrotor aircraft in dense indoor environments,'' \emph{Intl. J. Robot.
  Research ({IJRR})}, vol.~34, no.~7, pp. 969--1002, 2015.

\bibitem{zucker2013chomp}
M.~Zucker, N.~Ratliff, A.~D. Dragan, M.~Pivtoraiko, M.~Klingensmith, C.~M.
  Dellin, J.~A. Bagnell, and S.~S. Srinivasa, ``Chomp: Covariant hamiltonian
  optimization for motion planning,'' \emph{The International Journal of
  Robotics Research}, vol.~32, no. 9-10, pp. 1164--1193, 2013.

\bibitem{Gammell2015bit}
J.~D. {Gammell}, S.~S. {Srinivasa}, and T.~D. {Barfoot}, ``Batch informed trees
  (bit*): Sampling-based optimal planning via the heuristically guided search
  of implicit random geometric graphs,'' in \emph{2015 IEEE International
  Conference on Robotics and Automation (ICRA)}, 2015, pp. 3067--3074.

\bibitem{karaman2011}
S.~Karaman and E.~Frazzoli, ``Sampling-based algorithms for optimal motion
  planning,'' \emph{The International Journal of Robotics Research}, vol.~30,
  pp. 846--894, 2011.

\bibitem{liu2017ral}
S.~Liu, M.~Watterson, K.~Mohta, K.~Sun, S.~Bhattacharya, C.~J. Taylor, and
  V.~Kumar, ``Planning dynamically feasible trajectories for quadrotors using
  safe flight corridors in 3-d complex environments,'' \emph{IEEE Robotics and
  Automation Letters ({RA-L})}, pp. 1688--1695, 2017.

\bibitem{MelKum1105}
D.~Mellinger and V.~Kumar, ``Minimum snap trajectory generation and control for
  quadrotors,'' in \emph{Proc. of the {IEEE} Intl. Conf. on Robot. and Autom.},
  Shanghai, China, May 2011, pp. 2520--2525.

\bibitem{wang2020generating}
Z.~Wang, H.~Ye, C.~Xu, and F.~Gao, ``Generating large-scale trajectories
  efficiently using double descriptions of polynomials,'' \emph{arXiv preprint
  arXiv:2011.02662}, 2020.

\bibitem{Zhou2021EGO}
X.~{Zhou}, Z.~{Wang}, H.~{Ye}, C.~{Xu}, and F.~{Gao}, ``Ego-planner: An
  esdf-free gradient-based local planner for quadrotors,'' \emph{IEEE Robotics
  and Automation Letters}, vol.~6, no.~2, pp. 478--485, 2021.

\bibitem{Janson2015FMT}
L.~Janson, E.~Schmerling, A.~Clark, and M.~Pavone, ``Fast marching tree: A fast
  marching sampling-based method for optimal motion planning in many
  dimensions,'' \emph{The International Journal of Robotics Research}, vol.~34,
  no.~7, pp. 883--921, 2015.

\bibitem{nlopt}
\BIBentryALTinterwordspacing
S.~G. Johnson, ``The nlopt nonlinear-optimization package.'' [Online].
  Available: \url{http://github.com/stevengj/nlopt}
\BIBentrySTDinterwordspacing

\end{thebibliography}
\end{document}